\documentclass[10pt,twocolumn,letterpaper]{article}

\usepackage{iccv}
\usepackage{graphicx}
\usepackage{amsmath}
\usepackage{amssymb}
\usepackage{booktabs}
\usepackage{gensymb}
\usepackage{multirow}
\usepackage{rotating}
\usepackage{fancyhdr}
\usepackage{latexsym}
\usepackage{url}

\iccvfinalcopy 

\ificcvfinal\fi

\newlength\savewidth

\makeatletter\renewcommand\paragraph{\@startsection{paragraph}{4}{\z@}
  {.5em \@plus1ex \@minus.2ex}{-.5em}{\normalfont\normalsize\bfseries}}\makeatother

\usepackage{algorithm}
\usepackage{algorithmic}
\usepackage{bm}
\usepackage{stmaryrd}
\usepackage{dsfont}
\usepackage{epstopdf}
\usepackage{booktabs}
\usepackage{dashrule}
\usepackage{soul}
\usepackage{color}
\usepackage[table,x11names]{xcolor}

\usepackage{ragged2e}
\usepackage[pagebackref,breaklinks,colorlinks]{hyperref}

\usepackage[accsupp]{axessibility}

\newcommand{\blue}[1]{\textcolor{blue}{\bf{#1}}}
\newcommand{\red}[1]{\textcolor{red}{\bf{#1}}}
\makeatother
\usepackage{tikz}
\usepackage{pgfplots}
\usepackage{caption}
\usepackage{graphicx}
\usepackage{pifont}
\newcommand{\cmark}{\ding{51}}%

\newcommand{\kevin}[1]{\textcolor{black}{#1}}

\usepackage[capitalize]{cleveref}
\crefname{section}{Sec.}{Secs.}
\Crefname{section}{Section}{Sections}
\Crefname{table}{Table}{Tables}
\crefname{table}{Tab.}{Tabs.}

\def\iccvPaperID{5814} 

\DeclareUnicodeCharacter{2212}{-}

\begin{document}

\title{RCA-NOC: Relative Contrastive Alignment\\ for Novel Object Captioning}
%

\author{Jiashuo Fan $^2$ \quad
Yaoyuan Liang $^2$ \quad
Leyao Liu $^2$ \quad
Shaolun Huang $^2$ \quad
Lei Zhang$^{1}$ \quad \\
$^1$International Digital Economy Academy\quad
$^2$Tsinghua University\quad
}

\maketitle
\begin{abstract}
In this paper, we introduce a novel approach to novel object captioning which employs relative contrastive learning to learn visual and semantic alignment. Our approach maximizes compatibility between regions and object tags in a contrastive manner. To set up a proper contrastive learning objective, for each image, we augment tags by leveraging the relative nature of positive and negative pairs obtained from foundation models such as CLIP. We then use the rank of each augmented tag in a list as a relative relevance label to contrast each top-ranked tag with a set of lower-ranked tags. This learning objective encourages the top-ranked tags to be more compatible with their image and text context than lower-ranked tags, thus improving the discriminative ability of the learned multi-modality representation. We evaluate our approach on two datasets and show that our proposed RCA-NOC approach outperforms state-of-the-art methods by a large margin, demonstrating its effectiveness in improving vision-language representation for novel object captioning.

\end{abstract}

\section{Introduction}

Describing novel objects unseen in training data is a highly desired capability for a real-world image captioning model. Conventional image captioning models~\cite{Anderson2017up-down,vinyals2014neural,Lu2018nbt} often fail to describe novel objects because they only cover limited visual concepts and generalize poorly to images in the wild~\cite{tran2016rich}. To overcome this limitation, approaches that rely on object detection as the external resource\cite{venugopalan17noc,Lu2018nbt,wu2018decoupled,Berkan2019zsc,Mikihiro2020ove,xianyu2021anoc,li2020oscar,hu2021vivo,zhang2021vinvl} have been widely explored and demonstrated breakthroughs in vision-language (VL) understanding.  

\begin{figure}[t]
	\centering
	\includegraphics[width=\linewidth]{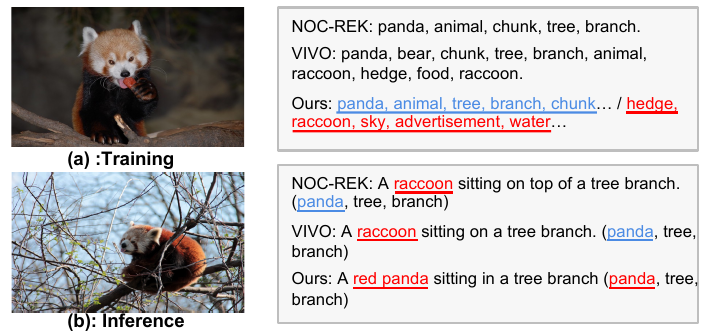}
        \captionsetup{font={small}}
	\caption{An illustrative example of our method to leverage relative semantic relevance to achieve modality alignment. Our method could give accurate captions \texttt{"A cute red panda sitting in a tree branch"} conditioned on the objects \texttt{"red panda"} while VINVL+VIVO generates a wrong caption \texttt{"A cute raccoon sitting in a tree branch"}. This inference result shows that our method can differentiate some confusing objects and generate accurate captions for novel objects when the object detection is well aligned with other modalities.}
	
	\label{fig:discriminative}
	\vspace{-2ex}
\end{figure}

Although object detection models (\textit{e.g.}, Faster RCNN~\cite{ren2015faster}) have been improved to recognize a wide range of objects including novel ones like zero-shot object detection~\cite{joseph2021towards}, using object detection in
novel captioning models brings a new challenge. VIVO~\cite{hu2021vivo} leverages extra object tags to pre-train a visual vocabulary and help image captioning generalize to new categories. NOC-REK~\cite{noc-rek} tries to augment object tags in the training stage based on the similarity between region and word. Such methods simply concatenate regions, object tags, and caption features as input to a Transformer-based model and use masked token reconstruction to implicitly learn an alignment between vision and language. Although such an alignment can help bring regions and words of the same concept closer, it lacks an effective mechanism to push away irrelevant concepts. As a result, it still makes mistakes on some confusing concepts, as shown in Figure \ref{fig:discriminative}.

Different from previous methods, this study aims to learn visual and semantic alignment in a contrastive manner. Gupta \etal \cite{gupta2020contrastive} showed that modality alignment could be achieved by maximizing the information lower bound between an image and its object tags. That is, given pairs of image and object tags, we can maximize the compatibility between tags and their attention-weighted region representations, compared to regions and non-corresponding tags. However, there are two critical problems that need to be further addressed: 1) how to effectively generate contrastive tags (augmented tags) that are closely relevant to an image, and 2) how to design a proper contrastive learning objective that allows the model to effectively leverage the contrastive tags to align vision and semantics.

To tackle the first problem, we utilize CLIP~\cite{clip} to create a list of contrastive tags which are closely linked to an image and contain global structural information and high-level concepts describing scenes. This approach aids in the discovery of useful tags that are essential for contrastive learning. To address the second problem, a proper contrastive learning objective needs to be explored. The major challenge here is that the augmented contrastive tags are inaccurate and inevitably noisy. We cannot simply treat one tag as positive and others as negative to perform contrastive learning, because the augmented tags might be highly correlated or similar to each other. To tackle this problem, we leverage the relative relevance, rather than the absolute relevance, of the augmented tags, which is more robust to data noise.

Specifically, given an image, for each of its labeled object tags, we generate a ranked list of contrastive tags using CLIP. We regard the rank of each augmented tag as its relative semantic relevance with the image. In general, the top-ranked tags are assumed to be more relevant to the image than the lower-ranked tags. We divide the list into two parts: the (relatively) relevant part with higher rank and the (relatively) irrelevant part with lower rank (Figure \ref{fig:discriminative} (a)). In our proposed objective, we treat each tag in the relevant part as positive and all tags in the irrelevant part as negative to perform contrastive learning. Note that we do not let tags both in the relevant part contrast with each other as they might be highly correlated concepts.
In this way, our approach weakens the strict assumption of contrastive learning in previous works and exploits the relative ranking in a loose form to achieve modality alignment.

We conduct experiments on the Nocaps and Held-Out COCO datasets to demonstrate the effectiveness of RCA-NOC. Our contributions can be summarized as follows:



$\bullet$ We propose Relative Contrastive Alignment (RCA) to learn the relative semantic relevance in a loose form by maximizing the compatibility between regions and their relevant tags compared with regions and irrelevant tags to achieve vision and language alignment and improve the discriminative ability of the multi-modality representation.

$\bullet$ A method called Uncertainty-Aware Selection and Reweighting (UASR) is proposed to estimate and exploit the uncertainty of each contrastive sample to mitigate the negative effect brought by noisy tags. UASR can effectively prioritize highly reliable samples and demote false positives and false negatives.

$\bullet$ We validate the proposed method on the Nocaps and Held-Out COCO benchmarks, which outperforms other state-of-the-art methods by a large margin.
\section{Related Work}
\label{sec:intro}

\textbf{Novel object captioning} aims to describe
images with objects that are unseen in the training stage (we define these objects as novel objects) where many methods.~\cite{venugopalan17noc,Lu2018nbt,wu2018decoupled,Berkan2019zsc,xianyu2021anoc,li2020oscar,hu2021vivo,zhang2021vinvl,clip,wang2021simvlm} have been proposed. Early works such as Hendricks \etal~\cite{hendricks16dcc} and Venugopalan \etal~\cite{venugopalan17noc} utilize unpaired labeled image and sentence data to enhance semantically visual concepts. Recent studies propose to explicitly leverage the object detection results for NOC,  Lu \etal~\cite{Lu2018nbt}, Wu \etal~\cite{wu2018decoupled}, and Demirel \etal~\cite{Berkan2019zsc} fill the generated template sentence with objects detected by object/novel object detectors. Chen \etal~\cite{xianyu2021anoc} combine object detector and human attention to identify novel objects. In addition, Li \etal are the first to utilize semantics in VLP tasks, which are further extended by Zhang \etal~\cite{zhang2021vinvl}. Hu \etal~\cite{hu2021vivo} built upon \cite{zhang2021vinvl} and propose to leverage extra region-tag pairs to conduct pretraining. Duc~\etal~\cite{noc-rek} tried to augment object tags in the training stage based on the similarity of regions and objects.


However, most aforementioned methods for NOC ignore the misalignment problem of object tags, thereby failing to fully exploit the semantic relationship between vision and language (Figure \ref{fig:discriminative}), which we argue is crucial to the quality of generated captions. In this paper, we propose a simple but effective contrastive learning objective to learn the relative semantic relevance in a loose form where the object tags could be explicitly aligned with their corresponding image feature representations in a semantic space.

\begin{figure*}[t]
	\centering	\includegraphics[width=\linewidth]{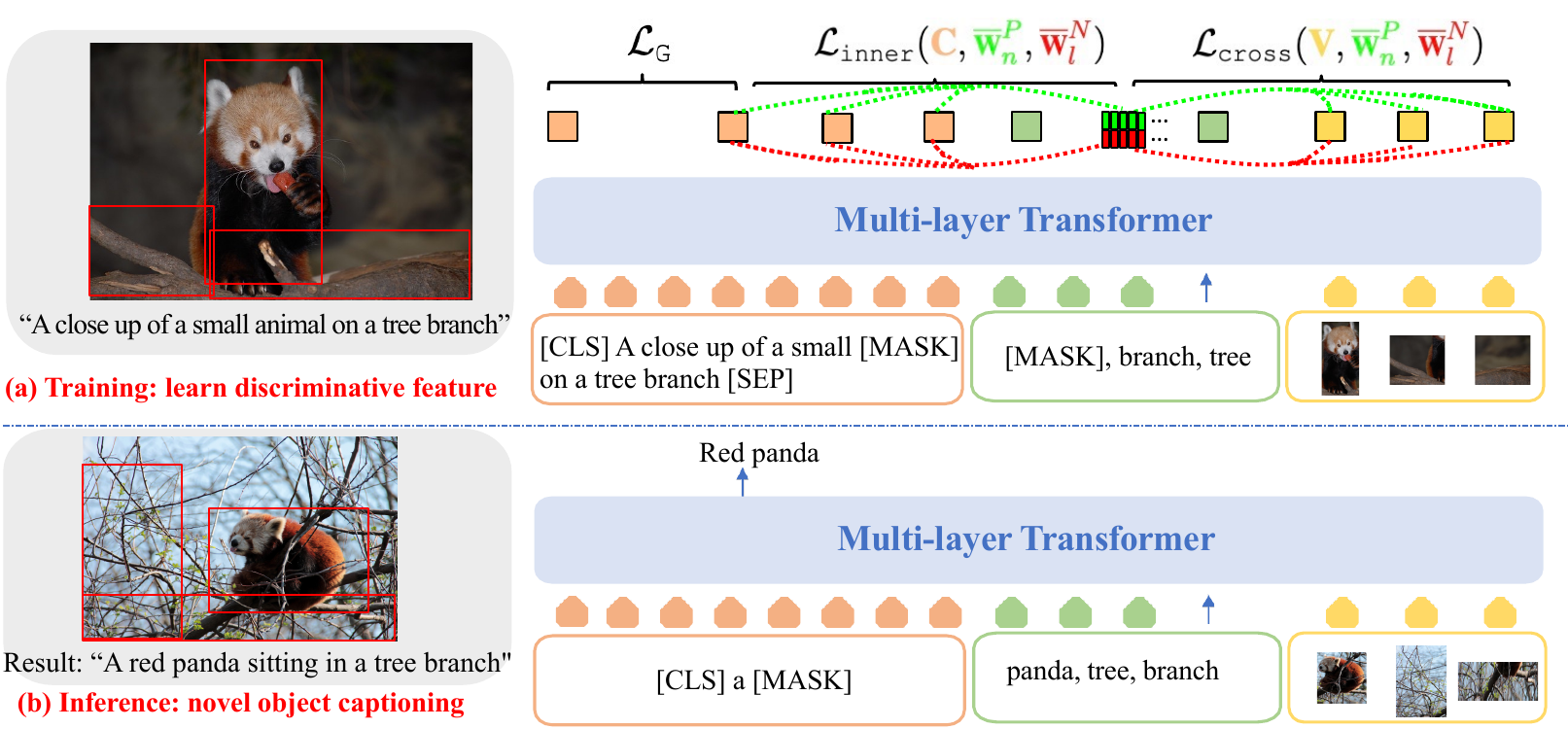}
        \captionsetup{font={small}}
	\caption{The main pipeline of our proposed RCA-NOC, including training and inference: In addition to the normal caption loss $L_G$ in the training process, we also compute the cross-modality and inner-modality loss $L_{cross}$ and $L_{inner}$ in Eq.~\ref{eq:cross1} and Eq.~\ref{eq:inner1} by contrasting the positive tags against the negative tags. The corresponding relationship is denoted in green and red line separately. During training, the input is a concatenated sequence of words—ROI tags—ROIs and augmented contrastive tags. The training loss comprises caption loss (with [MASK] in words for reconstruction) and contrastive learning loss (with [MASK] in ROI tags token for indicating the specific positions where augmented tags are inserted). Augmented tags replace certain ROI tags and are encoded into contrastive tag feature embeddings for comparison with region and caption embeddings. During prediction, the input is a concatenated sequence of ROI tags—ROIs without additional retrieval steps.}
	\label{fig:framework2}
	\vspace{-2ex}
\end{figure*}

\textbf{Contrastive Learning} aims to learn discriminative representations to distinguish an image from others. Many methods~\cite{chen2020simple,he2019momentum,wei2021aligning,licontextual,gupta2020contrastive,clip,li2021align} have shown their effectiveness. For example, Chen \etal\ \cite{chen2020simple} proposed to learn visual representations by maximizing agreement between differently augmented
views of the same image via a contrastive loss. He \etal\ \cite{he2019momentum} proposed Momentum Contrast (MoCo) for unsupervised visual representation learning. Wei \etal\ \cite{wei2021aligning} utilized contrastive learning in the detection task. Li~\etal~\cite{licontextual} focused on the problem of contextual outpainting. Gupta \etal~\cite{gupta2020contrastive} proposed a method CPG\footnote[1]{For better illustration, we use the abbreviation CPG to denote the paper "Contrastive Learning for Weakly Supervised
Phrase Grounding"~\cite{gupta2020contrastive}} to find hard negative words by replacing a word in a caption.

Other data-driven studies such as CLIP~\cite{clip} and ALIGN~\cite{li2021align} focus on learning a corresponding relative relationship from massive web data. CLIP~\cite{clip} predicts which text goes with which image and learns a relative image-text corresponding relationship from broad web data with noisy supervision. ALIGN~\cite{li2021align} further scales up CLIP by leveraging a noisy dataset that covers more than one billion image-text pairs. These methods achieve remarkable results but require massive fully-annotated data, which are difficult to obtain.

\section{Proposed Methods}

We propose to enhance the modality alignment by explicitly injecting visual semantics. These semantics are extracted for each image, obtained with existing foundation models such as CLIP. 

There are two motivations behind our approach. First, by maximizing the compatibility between regions and their relevant tags compared with regions and irrelevant tags, we could not only learn vision-semantic alignment and improve the discriminative ability of the multi-modality representation, Second, we introduce a relative contrastive learning objective that considers the relative relationships between positive and negative examples, rather than using absolute prototype-contrastive learning methods. This approach is more generalized and can be easily integrated into any existing NOC method.

\subsection{Visual Semantics Extraction}
 \label{sec:KD}

Different from other approaches (e.g., \cite{li2020oscar,hu2021vivo}) that extract object tags using pretrained object detectors, we use an off-the-shelf foundational model to extract more diverse, larger and semantically meaningful set of visual semantics. We show in section \ref{sec:Experiments}, that this approach helps to capture high level and global semantics describing the scenes, which are hard to obtain with other approaches (e.g. object detectors). 

We use a pretrained CLIP (ViT-B/16) model to obtain the embeddings of all the images and the extracted semantics. For each image, we compute its cosine similarity with all the embedded semantics and select the top M similar semantics as augmented tags {\footnotesize${T =\{t_l\}_{l=1}^{M} }$}. The augmented tags will be ranked using global (image-level) cosine similarity $p(t_l)$. Then we utilize the rank of each augmented tag in $T$ as a relative relevance label. In general, the top-ranked tags are assumed to be more relevant to the image than the lower-ranked tags. We divide $T$ into positive and negative tags: positive tags being the (relatively) relevant tags with higher rank {\footnotesize$T^P = \{{t_n^P}\}_{n=1}^{K} =\{t_l\}_{l=1}^{K}$} and negative tags being the (relatively) irrelevant tags with lower rank {\footnotesize$T^N = \{t_l^N\}_{l=1}^{K} =\{t_l\}_{l=K+1}^{2K}$ ($M=2K$)}.

Here we propose a simple yet effective augmentation technique. Having a set of visual semantics extracted for each image, instead of considering all the visual semantics at once, we sample randomly a fraction of these augmented tags at each iteration step. Hence, we could not only prevent the model from overfitting on specific semantics and potentially disregard the image or other semantics during training, but also let the model see different combinations of visual semantics, which helps to have more diversity.

\subsection{Relative Contrastive Alignment} 
\label{sec:CIAM}

InfoNCE~\cite{oord2018representation} is a typical type of contrastive loss function used for self-supervised learning that contrasts one positive sample against a set of negative samples. CPG~\cite{gupta2020contrastive} builds upon InfoNCE and proposes a compatibility function for regions and a caption word. Both InfoNCE and CPG utilize the absolute relevance to optimize the contrastive loss. 

However, such a contrastive learning objective is hard to optimize for object tags since they may be highly correlated to each other. In RCA-NOC, we focus on the relative semantic relevance and utilize the attention-based compatibility function to measure the relativity information for regions and a set of object tags, which is more robust to data noise.

Specifically, we first compute the dot product for a region-tag pair.

\vspace{-2ex}
\begin{equation}
s_{jk} = \textbf{w}_j^T \textbf{v}_k / \sqrt{d},
\end{equation}
where $\textbf{w}_j$ and $\textbf{v}_k$ refer to the corresponding embeddings for tag and region, and $d$ is the feature dimension. Here, $s_{jk}$ represents the similarity between the $j$-th tag and the $k$-th region. To find a contextualized region representation for the $j$-th tag, we define $\textbf{a}_j^v$ as follows.

\vspace{-2ex}
\begin{equation}
  \textbf{a}_{j}^{v}=\sum_{k=1}^K \alpha_{jk}\textbf{v}_k,
  \label{w_att}
\end{equation}
where 
\vspace{-3ex}
\begin{equation}
  \alpha_{jk} = \frac{e^{s_{jk}}}
  {\sum_{k'=1}^K e^{s_{jk'}}},
  \label{a}
\end{equation}

To measure the compatibility between a tag $\textbf{w}_j$ and its contextualized region representation $\textbf{a}_j^v$, we define

\begin{equation}
{\phi}(\textbf{V},\textbf{w}_j)= \textbf{w}_j^T{\textbf{a}_{j}^{v}}.
\label{compatibility}
\end{equation}

In this way, we can derive our cross-modality contrastive loss function $L_{cross}$. 

\vspace{-2ex}
\begin{small}
\begin{gather}
  \mathcal{L}_{\texttt{cross}}({\textbf{V},\textbf{W}^P,\textbf{W}^N}) = \notag\\
    -\frac{1}{K}\sum_{n=1}^{K}\log\left(
    \frac{e^{{\phi}(\textbf{V},\textbf{w}_n^P)}} {e^{{\phi}(\textbf{V},\textbf{w}_n^P)} + \sum_{l=1}^{K} e^{{\phi}(\textbf{V},\textbf{w}_l^N)}}
    \right),
\end{gather}\label{eq:cross}
\end{small}

\noindent where {\footnotesize$\textbf{W}^P=\{\textbf{w}_l^P\}_{l=1}^{K}$} and {\footnotesize$\textbf{W}^N=\{\textbf{w}_n^N\}_{n=1}^{K}$} are the word embeddings for  positives/negatives $T^P$ and $T^N$, extracted from CLIP. $\phi()$ is the compatibility function defined in Eq.~\ref{compatibility}.

This equation encourages each positive tag $\textbf{w}_n^P$ to be more compatible with image regions {\footnotesize$\textbf{V}$} than all negative tags in {\footnotesize$\textbf{W}^N$}. Essentially, if tag $t_j$ is not relevant to the image (negative), its representation $\textbf{w}_j$ should not be similar to its contextualized region representation $\textbf{a}_j^v$ since it would not be able to collect good information while computing $\textbf{a}_j^v$. Note that we do not let two positive tags (top-ranked tags) contrast with each other as it is hard to tell which one is more relevant.

\noindent\textbf{Inner modality Alignment.} To further enhance modality alignment, we also compute the inner-modality contrastive loss over tag-caption pairs. The formulation for inner-modality contrastive loss is similar to cross-modality contrastive loss but slightly different. We enforce contrastive learning not to be between tags and all caption words, but just between tags and noun caption words. This is because a caption may have many irrelevant tokens, such as "the, a, is", which will harm our alignment process and cause redundant computational costs.

Finally, we can derive our inner-modality contrastive loss function $L_{inner}$.

\vspace{-1ex}
\begin{equation}
{\phi}(\textbf{C},\textbf{w}_l)= \textbf{w}_l^T{\textbf{a}_{l}^{c}},
\label{eq:compatibility_inner}
\end{equation}
\vspace{-2ex}
\begin{small}
\begin{gather}
  \mathcal{L}_{\texttt{inner}}({\textbf{C},\textbf{W}^P,\textbf{W}^N}) = \notag \\
    -\frac{1}{K}\sum_{n=1}^{K}\log\left(
    \frac{e^{{\phi}(\textbf{C},\textbf{w}_n^P)}} {e^{{\phi}(\textbf{C},\textbf{w}_n^P)} + \sum_{l=1}^{K} e^{{\phi}(\textbf{C},\textbf{w}_l^N)}}
    \right),\label{eq:inner}
\end{gather}
\end{small}

\noindent where this time we take noun caption words ${\textbf{C}} = \{\textbf{c}_i\}_{i=1}^{P}$ in Eq.~\ref{eq:compatibility_inner} as input. By further considering inner-modality alignment, we not only learn caption-tag semantics to connect relevant caption tokens, but also enable region-caption interactions.

\subsection{Uncertainty-Aware Selection and Re-weighting} 
\label{sec:UASR}
The augmented contrastive tags are often noisy and would harm the discriminative ability and robustness of the contrastive learning process. Therefore, we design a sample selection strategy to deal with noisy contrastive tags (\textit{e.g.}, false positives/negatives).

\vspace{0.3ex}

\noindent\textbf{Uncertainty-Aware Selection.} Specifically, we first use the local cosine similarity to calculate the correlation between a region and a tag and filter false positives/negatives. The stronger the correlation, the more reliable a positive tag is and the more uncertain a negative tag. 
\vspace{-2ex}
\begin{equation}
    \operatorname{u}(\textbf{v}, \textbf{w}) = \frac{\textbf{v}^{\top} \textbf{w}}{\left \| \textbf{v} \right \| \left \| \textbf{w} \right \|},
    \label{eq:U1}
\end{equation}

Then we retrieve $L$ tags and their corresponding representations {\footnotesize${\textbf{H}}=\left \{{\textbf{w}_j}\right\}_{j=1}^{L}$} for the given $L$ region features {\footnotesize$\left \{ \mathbf{v}_i\right\}_{i=1}^{L}$}. We use $\mathop{\arg\max}$ to choose the most correlated tag $t_j$ for region $\textbf{v}_i$. 
\begin{equation}
    j = \operatorname*{arg \, max}_k (\operatorname{u}(\textbf{v}_i, \textbf{w}_k)),
\end{equation} 

According to the score in Eq.~\ref{eq:U1}, the corresponding top-L tags will be listed as confusion samples for negatives (actually positive), and these confusion samples will be removed from the original contrastive tags. Similarly, we will also choose the true positives based on these top-$L$ tags since tags with low-similarity scores should be regarded as outliers to the positives and they are too sparse to have a positive influence on the shape of embedding space. In this way, we can derive Eq.~\ref{eq:final_negative}, where $/$ is the removing set operation, $\cap$ is the intersection set operation. {\footnotesize$\overline{\textbf{W}}^P=\{\bar{\textbf{w}}_l^P\}_{l=1}^{K}$} and {\footnotesize$\overline{\textbf{W}}^N=\{\bar{\textbf{w}}_n^N\}_{n=1}^{K}$} is the final positive/negative tags' embedding sets. Note that if the set length of {\footnotesize$\overline{\textbf{W}}^P$}/{\footnotesize$\overline{\textbf{W}}^N$} is less than M, we will over-sample {\footnotesize$\overline{\textbf{W}}^P$}/{\footnotesize$\overline{\textbf{W}}^N$} until the set length is equal to M.

\vspace{-1ex}
\begin{small}
\begin{equation}
    \overline{\textbf{W}}^P=\textbf{W}^P\cap {\textbf{H}},\overline{\textbf{W}}^N=\textbf{W}^N/{\textbf{H}},
    \label{eq:final_negative}
\end{equation}
\end{small}
\noindent\textbf{Uncertainty-Aware Re-weighting.} To further mitigate the negative effect of noisy tags, we use the correlation function defined above to enhance reliable samples and reduce the influence of high-uncertainty samples. The information from more reliable samples should have a larger impact on the shape of the embedding and vice versa. Specifically, we first introduce the weight in Eq.~\ref{eq:r1}.

\vspace{-3ex}
\begin{gather}
        q_1(\bar{\textbf{w}}_n^P)=\exp{(\operatorname{u}(\textbf{v}_k, \bar{\textbf{w}}_n^P))}, \label{eq:r1}
\end{gather}

\vspace{-4ex}

\begin{gather}
        j = \operatorname*{arg \, max}_k (\operatorname{u}(\textbf{v}_i, \bar{\textbf{w}}_k^P)), \label{eq:r2}
\end{gather}

\vspace{-4ex}
\begin{gather}
        q_2(\bar{\textbf{w}}_n^P) = p(t_n),  \label{eq:r4}
\end{gather}

\noindent where $q_1(\bar{\textbf{w}}_n^P)$ in Eq.~\ref{eq:r1} is the uncertainty score computed by the most relevant region with the corresponding tag. Moreover, we further consider how certain each tag is and combine the corresponding score $q_2(\bar{\textbf{w}}_n^P)$ in Eq.~\ref{eq:r4} with $q_1(\bar{\textbf{w}}_n^P)$  to derive the final uncertainty score $q(\bar{\textbf{w}}_n^P)$ in Eq.~\ref{eq:r3}. Here $p(t_n)$ in Eq.~\ref{eq:r4} is the global similarity mentioned in Section~\ref{sec:KD}.

\vspace{-3ex}

\begin{gather}
        q(\bar{\textbf{w}}_n^P) = q_1(\bar{\textbf{w}}_n^P) * q_2(\bar{\textbf{w}}_n^P), \label{eq:r3}
\end{gather}

Thus, our method could mitigate the negative effect of noisy contrastive tags by considering the similarity from both the local region features (region-tag) and the
global structural information (image-tag). Finally, we could rewrite our contrastive loss function in Eq.~\ref{eq:cross1}, Eq.~\ref{eq:inner1}.
\begin{small}

\vspace{-1ex}
\begin{gather}
  \mathcal{L}_{\texttt{cross}}({\textbf{V},\overline{\textbf{W}}^P,\overline{\textbf{W}}^N}) = \notag\\
    -\frac{1}{K}\sum_{n=1}^{K}-q(\bar{\textbf{w}}_n^P)
    \log\left(
    \frac{e^{{\phi}(\textbf{V},\bar{\textbf{w}}_n^P)}} {e^{{\phi}(\textbf{V},\bar{\textbf{w}}_n^P)} + \sum_{l=1}^{K} e^{{\phi}(\textbf{V},\bar{\textbf{w}}_l^N)}}
    \right), ~\label{eq:cross1}
\end{gather}
\vspace{-3ex}

\begin{gather}
  \mathcal{L}_{\texttt{inner}}({\textbf{C},\overline{\textbf{W}}^P,\overline{\textbf{W}}^N}) = \notag\\
    -\frac{1}{K}\sum_{n=1}^{K}-q(\bar{\textbf{w}}_n^P)\log\left(
    \frac{e^{{\phi}(\textbf{C},\bar{\textbf{w}}_n^P)}} {e^{{\phi}(\textbf{C},\bar{\textbf{w}}_n^P)} + \sum_{l=1}^{K} e^{{\phi}(\textbf{C},\bar{\textbf{w}}_l^N)}}
    \right), ~\label{eq:inner1}
\end{gather}
\end{small}
 
\vspace{-3ex}


\renewcommand{\arraystretch}{1.5}
\begin{table*}[tb]
{\centering
\setlength{\tabcolsep}{7.5pt}
\captionsetup{font={small}}
\caption{Our evaluation results using SPICE and CIDEr on the Nocaps validation and test sets. We achieve the best scores for in-domain, near-domain, out-domain and Overall. Notably, the captions generated by our method are better than those by human in most cases. We note that our results on the test set are better than those by other methods which are publicly submitted to Nocaps leader-board$^b$. Higher score is better.}
	\vspace*{-0.25\baselineskip}
\label{tab:nocaps}
\resizebox{\linewidth}{!}{
\begin{tabular}{l|cc|cc|cc|cc|cc|cc|cc|cc}
\toprule
\multirow{3}{*}{Method} & \multicolumn{8}{c|}{Validation set} & \multicolumn{8}{c}{Test set} \\ 
& \multicolumn{2}{c|}{in-domain} & \multicolumn{2}{c|}{near-domain} & \multicolumn{2}{c|}{out-domain} & \multicolumn{2}{c|}{Overall} & \multicolumn{2}{c|}{in-domain} & \multicolumn{2}{c|}{near-domain} & \multicolumn{2}{c|}{out-domain} & \multicolumn{2}{c}{Overall}\\
& CIDEr& SPICE & CIDEr& SPICE & CIDEr& SPICE & CIDEr& SPICE & CIDEr& SPICE & CIDEr& SPICE & CIDEr& SPICE & CIDEr& SPICE\\
\hline
UpDown~\cite{Anderson2017up-down} & 78.1 & 11.6 & 57.7 & 10.3 & 31.3 & 8.3 & 55.3 & 10.1 & 76.0 & 11.8 & 74.2 & 11.5 & 66.7 & 9.7 & 73.1 & 11.2  \\
Oscar~\cite{li2020oscar} & 83.4 & 12.0 & 81.6 & 12.0 & 77.6 & 10.6 & 81.1 & 11.7 & 81.3 & 11.9 & 79.6 & 11.9 & 73.6 & 10.6 & 78.8 & 11.7 \\
VIVO~\cite{hu2021vivo} & 92.2 & 12.9 & 87.8 & 12.6 & 87.5 & 11.5 & 88.3 & 12.4 & 89.0 & 12.9 & 87.8 & 12.6 & 80.1 & 11.1 & 86.6 & 12.4 \\
VinVL~\cite{zhang2021vinvl} & 96.8 & 13.5 & 90.7 & 13.1 & 87.4 & 11.6 & 90.9 & 12.8 & 93.8 & 13.3 & 89.0 & 12.8 & 66.1 & 10.9 & 85.5 & 12.5 \\
VinVL + VIVO~\cite{zhang2021vinvl,hu2021vivo} & 103.7 & 13.7 & 95.6 & 13.4 & 83.8 & 11.9 & 94.3 & 13.1 & 98.0 & 13.6 & 95.2 & 13.4 & \red{78.0} & 11.5 & 92.5 & 13.1 \\
NOC-REK~\cite{noc-rek} & \red{104.7} & \red{14.8} & \red{100.2} & \red{14.1} & \red{100.7} & \red{13.0} & \red{100.9} & \red{14.0} &  \red{100.0} & \red{14.1} & \red{95.7} & \red{13.6} & 77.4 & \red{11.6} & \red{93.0} & \red{13.4} \\
\hline
RCA-NOC** & 95.0 & 13.6 & 91.3 & 13.1 & 93.2 & 12.1 & 92.2 & 13.0 & 87.7 & 13.3 & 88.1 & 12.9 & 77.9 & 11.4 & 86.3 & 12.7 \\
RCA-NOC & \blue{107.8} & \blue{15.3} & \blue{104.0} & \blue{14.6} & \blue{105.8} & \blue{13.6} & \blue{107.1} & \blue{14.6} &  \blue{104.1} & \blue{14.8} & \blue{101.2} & \blue{14.6} & \blue{88.5} & \blue{12.9} & \blue{101.1} & \blue{14.0} \\
\hline
$\Delta$ & 3.1$\uparrow$ & 0.5$\uparrow$ & 4.2$\uparrow$ & 0.5$\uparrow$ &5.1$\uparrow$ & 0.6$\uparrow$ & 6.2$\uparrow$ & 0.6$\uparrow$ & 4.1$\uparrow$ & 0.7$\uparrow$ & 5.5$\uparrow$ & 1.0$\uparrow$ & 11.1$\uparrow$ & 1.3$\uparrow$ & 8.1$\uparrow$ & 0.6$\uparrow$ \\
\hline
Human~\cite{agrawal2019nocaps} & 84.4 & 14.3 & 85.0 & 14.3 & 95.7 & 14.0 & 87.1 & 14.2 & 80.6 & 15.0 & 84.6 & 14.7 & 91.6 & 14.2 & 85.3 & 14.6 \\
\bottomrule
\end{tabular}
}
}
\vspace*{-1.25\baselineskip}
\end{table*}



\section{Experiments}
\label{sec:Experiments}


\subsection{Experimental Setup}
\noindent\textbf{Datasets.} Our main experiments and ablation studies are based on the Nocaps\cite{agrawal2019nocaps} dataset. Our method was built with PyTorch, and we used a pre-trained BERT-base model from Huggingfaces for parameters initialization, and no ground-truth tags are used on the Nocaps validation and test sets. For the training stage, we use the COCO training set which consists of 118K images, each with 5 captions. We evaluate our model on the validation and test sets of the Nocaps dataset, which consist of 4.5K and 10.6K images from the Open Images validation and test sets, respectively. Additionally, we test the proposed method on the Held-Out COCO \cite{hendricks16dcc}, which is a subset of MS COCO \cite{lin2014coco} where the following eight object categories are excluded from the training set: bottle, bus, couch, microwave, pizza, racket, suitcase, and zebra. We randomly split the COCO validation set and use half of it for validation and the other half for testing, each with 20,252 images. In addition, we empirically set $M=50$ to generate augmented object tags for the best performance.

\noindent\textbf{Implementation Details.}  In the training stage, the model is trained for 30 epochs with a batch size of 512 and a learning rate of $10^{−4}$, optimized using the cross-entropy loss and our contrastive loss. We set the maximum caption length to 40 and the maximum tag length to 30. To further boost the performance, we also perform the SCST optimization (Rennie \etal \cite{rennie2017self}) with a learning rate of 1.4 × $10^{−6}$ for 10 epochs. During inference, we use greedy decoding to generate image captions with a maximum length of 20. Our model is trained with 8 A100 GPUS and takes 1 day to train.

\renewcommand{\arraystretch}{0.9}
\begin{table}[tb]
\centering
\small
\captionsetup{font={small}}
\caption{Comparison of F1-scores (in \%) on object classes of Open Images, evaluated on the Nocaps validation set. There are $504$ classes in total. 80 of them are in-domain, which are common classes from COCO. The remaining $424$ classes are the out-of-domain objects.}
\vspace{-1mm}
\begin{tabular}{l@{\hspace{2mm}}|c@{\hspace{2mm}}c@{\hspace{2mm}}c}
\toprule
model  & in-domain & out-of-domain & entire  \\ \midrule
VIVO &  39.2 &  29.1 &  30.4 \\ NOC-REK & 45.3 &  30.5 &  32.8\\ 
Ours & 58.6 & 39.2 & 43.1 \\ 
\bottomrule
\end{tabular}
\vspace{-4mm}
\label{tab:nocaps_f1}
\end{table}

\subsection{Quantitative evaluation.} 
\noindent \textbf{Results on Nocaps dataset.} In this section, we extensively compare our frameworks with previous methods on the Nocaps benchmark. The other compared state-of-art results are from NOC-REK~\cite{noc-rek}. Note that all the
methods use the BERT-base\cite{devlin-etal-2019-bert} for a fair comparison. We compare our method with UpDown~\cite{anderson2018bottom,agrawal2019nocaps}, OSCAR~\cite{li2020oscar}, VinVL~\cite{zhang2021vinvl}, VIVO~\cite{hu2021vivo}, and NOC-REK~\cite{noc-rek}, which hold the state-of-the-art result on the Nocaps benchmark. The training data for the baselines is the COCO dataset. Following prior settings, we also report the results after the model is optimized using SCST~\cite{rennie2017self}. Our generated captions also adopt Constrained Beam Search (CBS) following ~\cite{anderson2016guided}.

In Table~\ref{tab:nocaps}, we present our results of SPICE and CIDEr scores on the Nocaps validation and test sets. We also show the results of our different training stages in the middle of the table, where RCA-NOC** denotes our first training stage result and RCA-NOC is our final result after CIDEr optimization. Note that we do not compare with the data-driven methods\cite{li2021align,wang2022git,yu2022coca} which use massive out-of-domain image-caption pairs in the training and fail to follow the rule of Nocaps. By leveraging relative semantic relevance in contrastive learning, our method has achieved a significant improvement compared to all prior works. It is worth noting that our first stage training could generate a comparable result with \cite{hu2021vivo,zhang2021vinvl,li2020oscar}. After the CIDEr optimization process, our method could outperform the recent state-of-art method NOC-REK~\cite{noc-rek} with a large margin, which is $8.1$ and $0.6$ (CIDEr and SPICE) better than \cite{noc-rek} on the Nocaps Test set. This suggests that our model is more capable of generating captions with novel objects.

To quantitatively evaluate how well the model can describe novel objects, we also calculate the F1-score between our generated and the ground-truth tags on the validation set. Table~\ref{tab:nocaps_f1} shows the comparison with VIVO and NOC-REK on the Nocaps validation set. We see that RCA-NOC improves NOC-REK and VIVO on F1-score substantially, especially for out-of-domain objects. This again verifies the effectiveness of RCA-NOC's discriminative ability to describe and distinguish novel objects.

\renewcommand{\arraystretch}{0.9}
\begin{table}[tb]
\centering
\small
\centering\setlength{\tabcolsep}{1pt}
\captionsetup{font={small}}
\caption{Evaluation results on the Held-Out COCO test set. The best
results are highlighted in red.}
\vspace{-2.8mm}
\begin{tabular}{l@{\hspace{2mm}}|c@{\hspace{2mm}}c@{\hspace{2mm}}c@{\hspace{2mm}}c}
\toprule
Method  & \kevin{Avg. F1-score} & \kevin{SPICE} & \kevin{Meteor} & \kevin{CIDEr}  \\ \midrule
NOC &  48.8 & -- & 21.4 & -- \\
NBT & 48.5 & 15.7 & 22.8 & 77.0\\
DNOC& 57.9 & -- & 21.6 & --\\
ZSC &  29.8 & 14.2 & 21.9 & --\\
ANOC&  64.3 & 18.2 & 25.2 & 94.7\\
VinVL+VIVO & 71.8 & 24.5 & 30.6 & 132.8\\
NOC-REK &\blue{76.3} & \blue{26.9} & \blue{32.8} & \blue{138.4}\\
\hline
RCA-NOC & \red{79.5} & \red{27.0} & \red{35.9} & \red{142.8}\\
\bottomrule
\end{tabular}
\vspace{-3.8mm}
\label{tab:cc_oid}
\end{table}


\noindent \textbf{Results on the Held-Out COCO dataset.} To further prove the generalization ability of
our method, we also conduct experiments on the Held-Out COCO dataset. As we can see from Table~\ref{tab:cc_oid}, our method consistently beats the baseline VinVL+VIVO with a large margin. The improvements are $7.7$, $2.5$, $5.3$, and $10.0$
with BERT-base on the F1, SPICE,
METEOR, and CIDEr metrics. Additionally, our method is superior to all other state-of-art methods. In particular, compared with the recent state-of-art method NOC-REK~\cite{noc-rek}, the recent state-of-art on the Held-Out COCO dataset, RCA-NOC achieves a $4.4$ improvement on CIDEr (from $138.4$ to $142.8$) and $3.1$ improvement on METEOR (from $32.8$ to $35.9$). The $3.2$ improvement on the F1 score also shows that RCA-NOC performs better in describing novel objects.

\vspace{1ex}
\subsection{Ablation study.}
\label{sec:ablation}
In this subsection, we will discuss every component's
contribution to our framework. If not mentioned by purpose, all methods are conducted on the Nocaps validation set with BERT-base. (we cannot use the Nocaps test set because only 5 times submissions are allowed for the test set).

\noindent\textbf{Effectiveness of visual semantics.} We enhance the modality alignment by incorporating visual semantics from foundational models, with a specific instantiation using CLIP. To evaluate the generalization ability of our approach, we modify the model to extract visual semantics and conduct further experiments to compare it with NOC-REK~\cite{noc-rek} and VIVO for generating visual semantics. Our baseline model is trained using a contrastive approach that ranks ROI tags extracted from Faster R-CNN based on their softmax values, distinguishing between relevant and irrelevant tags.

Our findings indicate that using NOC-REK to generate augmented tags provides only marginal improvement to our framework, resulting in a $0.8$ and $0.1$ (CIDEr and SPICE) increase compared to the baseline. We hypothesize that this limited improvement may be due to the restricted linguistic knowledge contained in BERT, which may not encompass a wide range of novel categories and lack higher-level and global concepts required to describe complex scenes. In contrast, utilizing a larger pool of noisy visual semantics (CLIP) extracted from captions leads to better results, with a $4.9$ and $0.4$ (CIDEr and SPICE) increase compared to the baseline. Using a smaller and cleaner pool of visual semantics extracted from the classes of various object detection datasets (VIVO) only results in a $2.0$ and $0.2$ (CIDEr and SPICE) increase compared to the baseline. This is likely due to the fact that the CLIP is pre-trained on vast amounts of training data, enabling it to extract a more diverse, extensive, and semantically meaningful set of high-level visual concepts compared to NOC-REK.

\begin{figure*}[tb]
	\centering
	\includegraphics[width=1\linewidth]{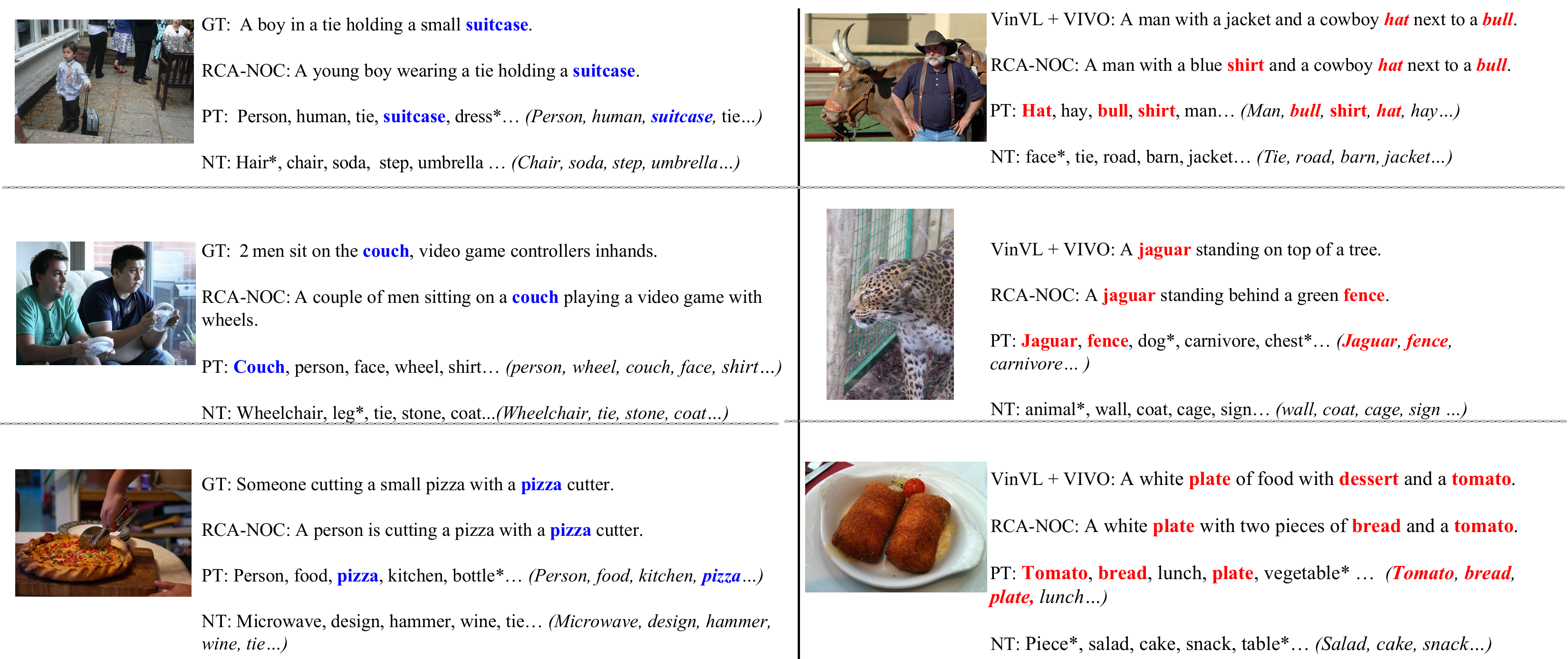}
    \caption{Examples of generated captions and contrastive tags by compared methods on Held-Out COCO (left) and Nocaps (right). We show the ground-truth captions (GT) on Held-Out COCO for reference. PT/NT denotes the positive/negative tags before and after Uncertainty-Aware Selection and Reweighting (UASR). \blue{Blue}/\red{Red} text indicates novel objects in Held-Out COCO/Nocaps, and $*$ indicates the false positives/negatives.}
	\label{fig:Contrastive_neg_samples}
	\vspace*{-1.5\baselineskip}
\end{figure*}


\begin{table}[tb]
\centering
\scriptsize
\renewcommand{\arraystretch}{1.1}
\centering\setlength{\tabcolsep}{0.4pt}
\captionsetup{font={small}}
\caption{Effectiveness of the source of visual semantics.} 
	\vspace*{0.25\baselineskip}
\label{tab:KDRM}
\vspace{-2.8mm}
\begin{tabular}{l|cc|cc|cc|cc}
\toprule
\multirow{2}{*}{Method} & \multicolumn{2}{c|}{in-domain} & \multicolumn{2}{c|}{near-domain} & \multicolumn{2}{c|}{out-of-domain} & \multicolumn{2}{c}{Overall}\\
& CIDEr& SPICE & CIDEr& SPICE & CIDEr& SPICE & CIDEr& SPICE \\
\midrule
N.A. & 103.6 & 14.7 & 100.8 & 14.2 & 100.9 & 13.2 & 103.2 & 14.2 \\
NOC-REK & 103.9 & 14.9 & 101.6 & 14.3 & 102.0 & 13.4 & 103.4 & 14.3\\
VIVO & 105.2 & 15.1 & 103.5 & 14.4 & 103.6 & 13.4 & 105.2 & 14.4 \\
CLIP & 107.8 & 15.3 & 104.0 & 14.6 & 105.8 & 13.6 & 107.1 & 14.6 \\
\bottomrule
\end{tabular}
\vspace*{-\baselineskip}
\end{table}

\begin{table}[tb]
\centering
\renewcommand{\arraystretch}{1.1}
\centering\setlength{\tabcolsep}{1pt}
\captionsetup{font={small}}
\caption{Effectiveness of the UASR. UAR: Uncertainty-Aware Re-weight, UAS: Uncertainty-Aware Selection, ${UAS}^N$: Uncertainty-Aware Selection for Negatives), UASR: Uncertainty-Aware Selection and Re-weight.} 
	\vspace*{0.25\baselineskip}
\label{tab:UASR}
\scriptsize
\vspace{-2.8mm}
\begin{tabular}{l|cc|cc|cc|cc}
\toprule
\multirow{2}{*}{Method} & \multicolumn{2}{c|}{in-domain} & \multicolumn{2}{c|}{near-domain} & \multicolumn{2}{c|}{out-of-domain} & \multicolumn{2}{c}{Overall}\\
& CIDEr& SPICE & CIDEr& SPICE & CIDEr& SPICE & CIDEr& SPICE \\
\midrule
N.A. & 103.6 & 14.7 & 100.8 & 14.2 & 100.9 & 13.2 & 102.4 & 14.2 \\
${UAS}^N$ & 105.2 & 14.8  & 102.1 & 14.3 & 102.4 & 13.3 & 104.2 & 14.3 \\
UAS & 105.9 & 15.0 & 104.5 & 14.3 & 103.9 & 13.4 & 105.3 & 14.5\\
UAR& 104.3 & 14.8 & 103.3 & 14.4 & 104.7 & 13.3 & 105.6 & 14.3 \\
UASR & 107.8 & 15.3 & 104.0 & 14.6 & 105.8 & 13.6 & 107.1 & 14.6 \\
\bottomrule
\end{tabular}
\vspace{-2.0mm}
\vspace*{-\baselineskip}
\end{table}

\begin{table}[tb]
\centering
\renewcommand{\arraystretch}{1.1}
\centering\setlength{\tabcolsep}{2pt}
\scriptsize
\label{tab:Karpathy}
\captionsetup{font={small}}
\caption{Evaluation on COCO test set of Karpathy split~\cite{karpathy2015deep}. All results are based on single model with cross-entropy optimization.}
\vspace{-2.8mm}
\begin{tabular}{l@{\hspace{6mm}}|c@{\hspace{6mm}}c@{\hspace{6mm}}c@{\hspace{6mm}}c}
\toprule
pre-training  & \kevin{BLEU4} & \kevin{Meteor} & \kevin{CIDEr} & \kevin{SPICE}  \\ \midrule
N.A & $33.8$ & $ 28.1$ & $118.3$ & $21.2$\\
OSCAR + VIVO & $ 34.9$ & $ 28.4$ & $ 119.8$ & $21.7$\\
RCA-NOC & $\bf 37.4$ & $\bf 29.6$ & $\bf 128.4$ & $\bf 23.1$\\
\bottomrule
\end{tabular}
\vspace{-2mm}
\label{tab:Karpathy}
\end{table}

\begin{table}[tb]
\centering\setlength{\tabcolsep}{10pt}
\renewcommand{\arraystretch}{1.1}
\captionsetup{font={small}}
\scriptsize
\caption{Ablation study on the effectiveness of different components, including (i) CA: Cross-modality Alignment,  our prototype contrastive loss for region-tag pairs, (ii) IA: Inner-modality Alignment, (iii) CLIP: Without CLIP refers to extracting visual semantics with object detection, and (iv) UASR: Uncertainty-Aware Selection and Re-weighting.} 
\label{tab:diffc}
\vspace{-2mm}
\label{tab:abla-of-losses}
\scriptsize
\begin{tabular}{cccc|cc}
\toprule
CA  & IA & UASR &  CLIP&  CIDER&  SPICE\\ \hline
  &   &  & &   $90.9$ & $12.8$\\ 
  \cmark  &  &  &  &  $96.5$ &  $13.5$\\\
   \cmark &  \cmark &  &  &  $98.7$  &$13.8$ \\
  \cmark  &\cmark &  \cmark & &$103.2$   &$14.2$  \\
  \cmark  & \cmark & \cmark &  \cmark & $107.1$   &$14.6$  \\ \bottomrule
\end{tabular}
\vspace{-4mm}
\vspace*{-\baselineskip}
\end{table}


\noindent\textbf{Effectiveness of the UASR.} We investigate the performance of our Uncertainty-Aware Selection and Reweighting (UASR) on the Nocaps validation set in this part. Table~\ref{tab:UASR} shows that Uncertainty-Aware Reweighting (UAR) could effectively enhance the reliable contrastive samples and bring $3.2$ and $0.1$ boosts on CIDEr and SPICE. In addition, the false positives/negatives severely harm our training process. By further incorporating Uncertainty-Aware Selection for Negatives (${UAS}^N$) into our method and filtering false negatives, our model's performance is boosted to $104.2$ and $14.3$ on CIDEr and SPICE. When filtering false positives/negatives simultaneously, we could reach $105.3$ and $14.5$ on CIDEr and SPICE with Uncertainty-Aware Selection (UAS). Finally, our method improves the performance by $4.7$ and $0.4$ (CIDEr and SPICE) by combining UAR and UAS, which illustrates the proposed UASR is a more powerful tool for tackling noisy contrastive samples in NOC.

\subsection{Qualitative Results}
In Figure \ref{fig:Contrastive_neg_samples}, we display some qualitative results on Held-Out COCO and Nocaps, and all the approaches are based on BERT-base. On Held-Out COCO, the result compared with GT shows that our proposed method could effectively generate accurate and precise captions with novel objects by leveraging relative semantic relevance into training. On Nocaps, VinVL+VIVO~\cite{zhang2021vinvl,hu2021vivo} sometimes cannot include the desired novel objects in their captions (first and third examples) or generate wrong captions (second example). Our RCA-NOC, on the other hand, successfully generates correct and coherent captions via differentiating confusing classes and aligning object detection with other modalities. For better illustration, we show top-5 Positive and Negative Tag results extracted from CLIP.

\subsection{General Image Captioning}
Improving modality alignment is a shared objective of general image captioning tasks. As demonstrated in Table~\ref{tab:Karpathy}, our proposed method, RCL-NOC, improves the model's performance across all metrics assessed on the COCO test set, particularly in the CIDEr score. However, we have observed that the improvement on the COCO benchmark is not as substantial as that on the nocaps benchmark. We hypothesize that this discrepancy is due to the COCO dataset having a limited number of visual concepts, thus reducing the benefits of learning visual semantics. Additionally, our method's use of tags is general and does not rely on fully annotated data. It's possible to utilize potentially unlimited amounts of images and different sources of tags, including those from Faster R-CNN, image tagging, or keywords extracted from captions. These possibilities will be explored in our future work..

\subsection{Data and Compute Efficiency}

The way the visual semantics are extracted (using CLIP) is not central to our work, as for the gain coming from 400M pairs of CLIP. This is supported in section \ref{sec:ablation}. Table~\ref{tab:diffc} shows that other components contribute significantly. Notably, the gain achieved by our Contrastive Alignment (CA) component alone can surpass that of CLIP by a large margin. Importantly, our method achieves this without increasing the number of parameters during training and inference, with the same parameter count as NOC-REK and VINVL (110 M). The training time of NOC-REK is 49 hours, for NOC-REK the training time is much lower (23h). This is with the paper setup (8 GPUs A100). Compared with NOC-REK, which adopts test time augmentation during inference, our inference time is much lower (1 minute) while NOC-REK is 2 minutes.

\section{Conclusion}

In this paper, we present RCA-NOC, which achieves visual-semantic alignment via relative contrastive learning. Specifically, for each image, we first extract augmented tags obtained from foundation models such as CLIP. Then we utilize the rank of each augmented object tag in a list as a relative relevance label to contrast each top ranked tag with a set of lower ranked tags. We empirically find that such a learning objective is effective and easy to optimize by encouraging the top ranked tags to be more compatible with their image and text context than lower ranked tags, hence improving the discriminative ability of the learned multi-modality representation. We prove the effectiveness of our paradigm in novel object caption, with the spotlight on Nocaps and Held-Out COCO benchmark.

The effectiveness of object tags has already been proved in recent works. However, there are few studies to further explore why such tags could perform well and the problem of object tags' misalignment is often ignored. So how to further exploit use tags explicitly while having a deep understanding of their roles is meaningful in the future.

\newpage





\end{document}